\documentclass[10pt,twocolumn,letterpaper]{article}

\usepackage{cvprstyle/cvpr}
\usepackage{times}
\usepackage{epsfig}
\usepackage{graphicx}
\usepackage{amsmath}
\usepackage{amssymb}

\usepackage{color}

\usepackage{graphicx}
\usepackage{amsmath,amssymb} 

\usepackage{longtable}
\usepackage{fancyhdr}
\usepackage[us,12hr]{datetime}
\usepackage{paralist}

\usepackage{tikz}

\DeclareMathOperator*{\argmin}{arg\,min}
\DeclareMathOperator*{\argmax}{arg\,max}

\usepackage[breaklinks=true,bookmarks=false]{hyperref}

\cvprfinalcopy 

\def\cvprPaperID{5379} 

\ifcvprfinal\fi
\begin{document}

\title{End-to-End Learned Random Walker\\ for Seeded Image Segmentation}

\author{
    Lorenzo Cerrone, Alexander Zeilmann, Fred A. Hamprecht \\
    Heidelberg Collaboratory for Image Processing \\
    IWR, Heidelberg University, Germany \\
    \small{
    \texttt{\{lorenzo.cerrone, alexander.zeilmann, fred.hamprecht\}@iwr.uni-heidelberg.de}}
}

\maketitle
\thispagestyle{empty}


\begin{abstract}
    We present an end-to-end learned algorithm for seeded segmentation.
    Our method is based on the Random Walker algorithm, where we
    predict the edge weights of the underlying graph using a convolutional
    neural network.
    This can be interpreted as learning context-dependent diffusivities for a
    linear diffusion process.
    Besides calculating the exact gradient for optimizing these diffusivities,
    we also propose simplifications that sparsely sample the gradient and still
    yield competitive results.
    The proposed method achieves the currently best results on a
    seeded version of the CREMI neuron segmentation challenge.
\end{abstract}



\section{Introduction}
Image segmentation is the task of partitioning an image into regions that are meaningful for a given task.
Seeded segmentation is a popular semi-supervised variant, where an oracle
annotates one or more pixels per region with an instance label,
such that all seeds in a region share the same label.

Most seeded segmentation pipelines involve two parts: a first step that predicts, for each pair of adjacent pixels, if or not they are likely part of the same region; and a second step that ingests these predictions as well as the seed locations and then infers the precise extent of each region. The second step is often cast as inference in an undirected probabilistic graphical model, either discrete~\cite{Boykov01} or continuous~\cite{Grady06}.

If no prior information on the nature of the images to be segmented is available, the first step -- boundary estimation -- is of necessity generic. It is then possible to use or train some state of the art edge detection method (e.g.~\cite{Maninis18}) and choose the parameters of the second step -- inference -- by hand.

But since dense ground truth is required for the training of a powerful edge detection method in any case, one may also go further and train the two steps {\it jointly}. Such a {\it structured} learning is more general than the naive approach of training a CNN to predict good boundary indicators, oblivious to what these will be used for; and to subsequently and independently optimize the parameters of the inference step.

For the second, inference, step, we choose a Conditional Gaussian Markov Random Field with only seed fidelity as the unary term -- also known as Random Walker algorithm~\cite{Grady06} -- and here show how to train that inference scheme jointly with a deep convolutional neural network (CNN).

Our choice of a Gaussian Markov Random Field is motivated by the following features:
\begin{itemize}
\item Less sensitive to noisy edge weights than the Watershed~\cite{Falcao04} which is fast and hence popular in seeded segmentation, but also susceptible to outliers based on its use of the (max, min)-semiring~\cite{Baras10}.
\item More amenable to differentiation than the purely combinatorial graph cuts problem~\cite{Boykov01}.
\item More computationally efficient than more expressive generalizations involving higher-order potentials~\cite{Roth05}.
\end{itemize}

More specifically, we make the following contributions:
\begin{enumerate}
    \item We demonstrate end-to-end structured learning of a pipeline consisting
        of a deep neural network that predicts edge weights, which are
        subsequently used in a seeded segmentation by linear diffusion, i.e., with the Random Walker
        algorithm.
    \item We calculate the exact gradient of our objective function and propose
        a sparse sampling strategy of the gradient to reduce computation time.
    \item We evaluate the algorithm on a seeded version of the MICCAI Challenge on Circuit
        Reconstruction from Electron Microscopy Images (CREMI)~\cite{CREMI}. Here, the proposed method outperforms unstructured training of a CNN combined with the Random Walker algorithm, and defines a new state of the art.
    \item We provide the source code as PyTorch package on \url{https://github.com/hci-unihd/pytorch-LearnedRandomWalker}, allowing general use of the proposed Learned Random Walker.
\end{enumerate}
\begin{figure*}
    \begin{center}
    \includegraphics[height=9cm]{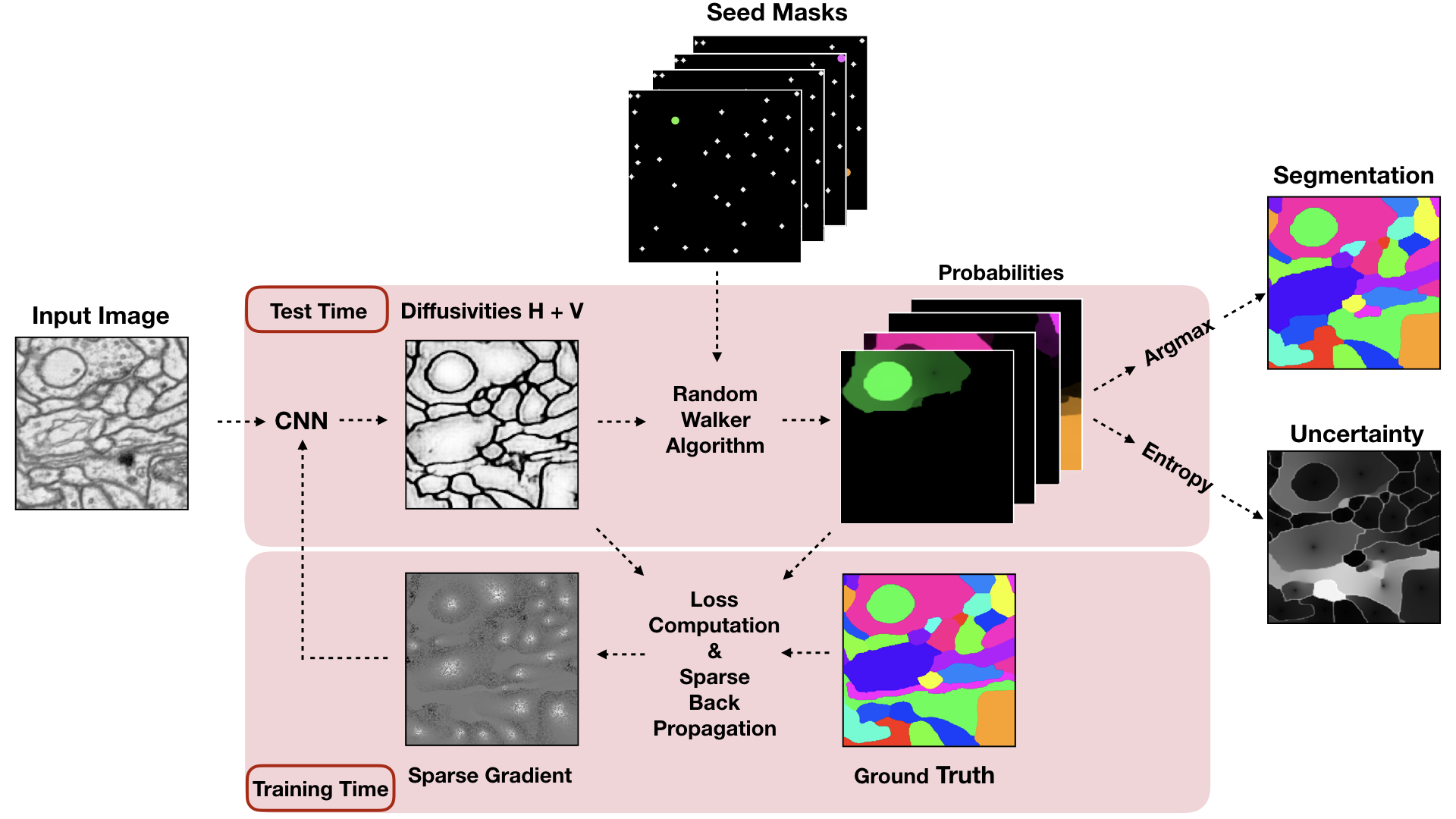}
    \end{center}
    \caption{Illustration of the Learned Random Walker pipeline at training and test time.
        \textbf{Top:} forward pass, the input image is mapped by a
        CNN to an undirected edge-weighted graph.
        A probability map is computed for each object using the Random Walker
        algorithm.
        At test time, a segmentation and an uncertainty map are computed from the
        probabilities.
        \textbf{Bottom:} backward pass of the training.
        The dense derivative of the Random Walker algorithm is too expensive to
        compute, instead, a sampling scheme is used and only a very sparse
        Jacobian is passed to the CNN.
    }\label{fig:masterfig}
\end{figure*}

\section{Related Work}

Diffusion processes are attractive because they engender nontrivial global behavior as a consequence of trivial local interactions. As such, they have a rich  history in machine learning and computer vision.

Seeded segmentation could be seen as an instance of semi-supervised segmentation, which has been explored in conjunction with Gaussian Markov Random Fields (GMRF) on arbitrary graphs in~\cite{Zhu03},
though without the supervised learning of edge weights as discussed here.

In computer vision, linear diffusion has been proposed for seeded segmentation in the seminal work of Grady~\cite{Grady06}.
In semantic segmentation and image-valued regression tasks, Tappen et al.~\cite{tappen09} have shown how to learn context-dependent potentials for GMRFs. These authors did not explore seeded segmentation, and the number of parameters learned was in the hundreds, not hundreds of thousands as in our case.

Jancsary et al.~\cite{Jancsary12} have pioneered the use of flexible nonparametric classifiers to determine the potentials in a Conditional GMRF. However, they opted for ensembles of decision trees and had to estimate the impact of each possible split on the loss function, making a number of approximations necessary.

In recent work, Vernaza and Chandraker~\cite{Vernaza} have proposed a first-order approximation of the
derivative for backpropagation. We show in section~\ref{sec:comparison} that this approximation is not well-behaved for the kind of sparse seeds considered here.

Our algorithm can also be seen from the viewpoint of coupling a deep network with an undirected probabilistic graphical model. In contrast to~\cite{chen2015,zheng2015} we use a GMRF that makes inference easier. End-to-end learning of a GMRF was also proposed in~\cite{Bertasius17}, though these authors do not solve the inference problem exactly as we do; instead, they unroll a few steps of gradient descent into extra layers of their neural network. In a similar setting, Chandra and Kokkinos ~\cite{Chandra2016} learned unary and pairwise terms of a Gaussian CRF, this time, computing the backpropagation derivative exactly.
Our inference problem can be seen as convex quadratic program. Recent work proposes quadratic programming as a generic neural network layer~\cite{optnet}. However, the paper and implementation do not account for the kind of sparsity that characterizes our formulation, and scales only to a few hundred random variables, where we use tens of thousands. Indeed, the solution of huge sparse systems of the kind discussed here is possible in almost linear time~\cite{Spielman04}.

Other important seeded segmentation strategies are based on discrete Markov Random Fields~\cite{Boykov01} that do however not admit the differentiability we crucially need; or on shortest paths similarities that can be computed even more efficiently than the diffusions we use~\cite{Sapiro07}.
However, the use of (single) shortest paths to measure the dissimilarity from each pixel to every seed is not as robust as the averaging over many paths that diffusion processes implicitly accomplish~\cite{vonLuxburg07}.

Such fragility is especially pronounced for watershed-type methods that base dissimilarities on a minimax criterion~\cite{Cousty10}.
Even so, such methods have been used successfully in biomedical~\cite{Malmberg2012} and other applications.
Recent work has sought to mitigate this limitation by learning the edge weights for watershed seeded segmentation in a structured fashion, just as we do, and moreover making the algorithm adaptive~\cite{LWS}, the cost for the latter being a somewhat involved recurrent neural net formulation. The experiments in section~\ref{sec:results} show that we can supersede the learned watershed in spite of our simpler architecture.

Remarkably, all of the above inference schemes for seeded segmentation -- graph cuts, random walker / linear diffusion and geodesic distances --  emerge as special cases of a unifying optimization problem~\cite{PWS}.

\section{Methods}\label{sec:math_background}
\subsection{Mathematical Background}
To make this paper self-contained, we derive the linear
system~\eqref{eq:RandomWalkerLinearSystem}
which is solved in the Random Walker algorithm.
While Grady~\cite{Grady06} deduced the system by minimizing a discretized
Dirichlet energy, we derive it directly from the discretized Laplace equation.

The Random Walker algorithm models seeded segmentation with $|\mathcal{L}|$ distinct categories of seeds as follows:
each category or label $a\in\mathcal{L}$ is associated with an infinitely large reservoir. This reservoir is coupled to the pixels that have been marked with label $a$. From this reservoir, the label $a$ can diffuse across the image. The pixels associated with all other labels $a'\in \mathcal L \setminus \{a\}$ act as sinks for label $a$. That is, the concentration of a label $a$ is one at the pixels that have been marked with this label; it is zero at the locations of all other seeds; and its ``concentration'' in-between is given by the stationary state of the linear diffusion equation. Importantly, local diffusivities are informed by local image appearance: Random Walker assumes high diffusivity within ground truth segments and low diffusivity across segments.

In the continuous setting, diffusion is usually formulated as
\begin{alignat*}{1}
	u &= f \text{ on } \partial \Omega, \\
	\Delta u &= 0 \text{ in } \Omega,
\end{alignat*}
where we prescribe the value of a function $u$ on the boundary of the domain
$\Omega$ and require that the Laplace operator vanishes in the interior of the
domain.

To carry these ideas over to the domain of discrete images, we consider an image as a connected undirected edge-weighted graph $G=(V, E)$ with adjacency relation
$i \sim j \Longleftrightarrow (i,j) \in E$ and edge weights
$w_e \in \mathbb{R}_+$ for every edge $e \in E$.
The set of vertices $V$ is partitioned into unmarked vertices $U$ and marked
vertices $M$, the seed pixels.
By $L$ we denote the graph Laplacian which is defined as
\begin{equation*}
	L_{i, j} =
	\begin{cases}
		- w_{i,j}                 & \text{if } i \sim j\\
		\sum_{k \sim i} w_{i, k}  & \text{if } i = j \\
		0                         & \text{else},
	\end{cases}
\end{equation*}
i.e.\ $L = D - A$ with the degree matrix $D$ and the adjacency matrix $A$ of $G$.
We consider the vertices ordered in such a way that the marked pixels appear
above and left of the unmarked pixels in $L$:
\begin{equation*}
	L = \begin{pmatrix}
		L_M & B \\
		B^T & L_U
	\end{pmatrix}.
\end{equation*}

We define a row stochastic matrix $Z \in {[0,1]}^{|V| \times |\mathcal{L}|}$ by
\begin{equation*}
    Z_{i,a} = \text{probability of vertex $i$ having label $a \in \mathcal{L}$}. 
\end{equation*}
This matrix $Z$ is called assignment matrix in recent
literature~\cite{Astroem2016}.
We assume that the rows in $Z$ are sorted in the same way as in $L$, i.e.\ the
marked rows are above the unmarked rows.
Thus, we can partition the matrix into a marked (known) part $Z_M$ and an unmarked (wanted) part $Z_U$.
With these notions, the continuous diffusion equation can be extended to the discrete setting in the following way:
\begin{equation*}
	\begin{pmatrix}
		L_M & B \\
		B^T & L_U
	\end{pmatrix}
	\begin{pmatrix}
		Z_M \\
		Z_U
	\end{pmatrix}
	=
	\begin{pmatrix}
		* \\
		0
	\end{pmatrix},
\end{equation*}
i.e.\ the Laplace matrix multiplied by the assignment matrix is $0$ on the
unmarked vertices and not prescribed on the marked ones --- indicated by the $*$
on the right-hand side.
This is similar to $\Delta u$ not being required to have a certain value on
$\partial \Omega$.
$Z_M$ is set to user-specified values.
Multiplying out yields the following extremely sparse linear system, which is at the heart of the
Random Walker algorithm:
\begin{equation}\label{eq:RandomWalkerLinearSystem}
	L_U Z_{U} = - B^T Z_{M}.
\end{equation}
Since $L_U$ is invertible, the solution $Z_U$ of the linear system exists
and is unique.

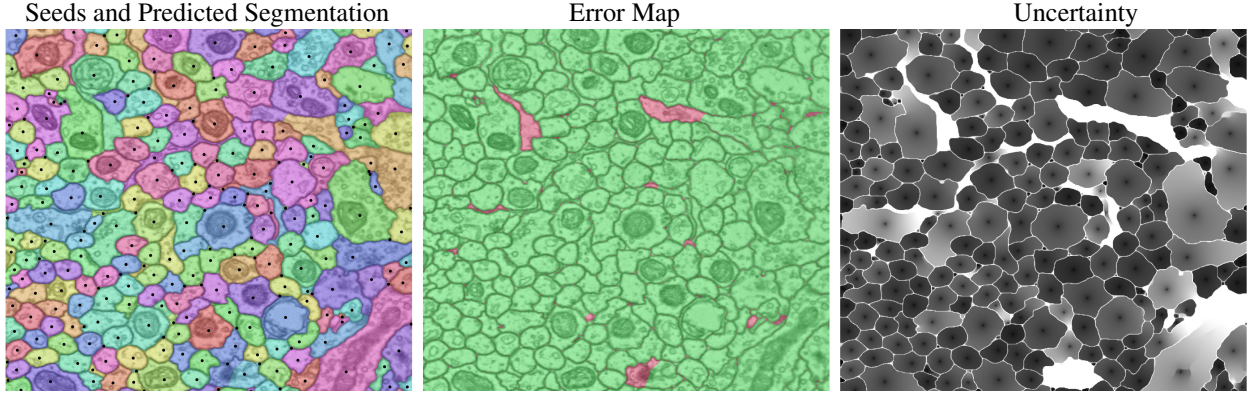
\begin{figure*}
    \centering
    \begin{tikzpicture}
        \node (img) {
            \includegraphics[width=0.95\textwidth]{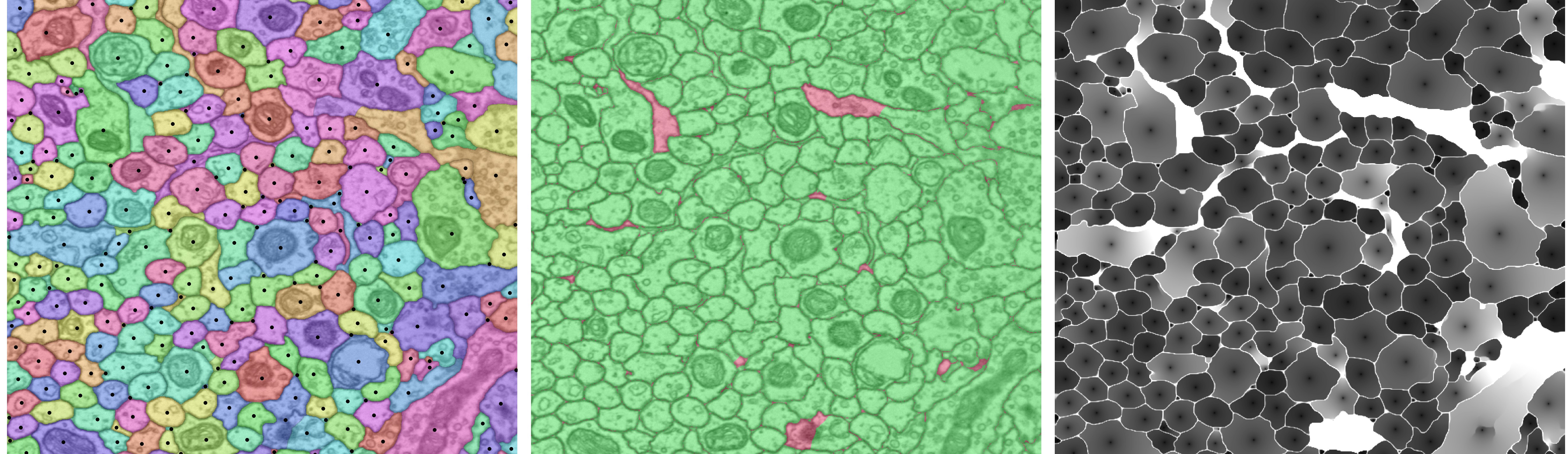}};
        \node[above of =  img, node distance=0cm, xshift=-5.55cm, yshift=+2.6cm,font=\color{black}] {Seeds and Predicted Segmentation};
        \node[above of =  img, node distance=0cm, yshift=+2.6cm,font=\color{black}] {Error Map};
        \node[above of =  img, node distance=0cm, xshift=6.0cm, yshift=+2.6cm,font=\color{black}] {Uncertainty};
    \end{tikzpicture}
    \caption{The assignment matrix produced by the Random Walker algorithm can
        be converted to a segmentation (left picture, including the seeds we
        used in the Random Walker algorithm).
        The center image illustrates the wrongly labeled pixels (marked red).
        The uncertainty of the labeling is illustrated via the entropy (white
        indicates high, black indicates small uncertainty).\newline
        We found that wrongly segmented regions usually have high uncertainty. However the
        converse is not true in general: High uncertainty does not necessarily indicate a
        wrong segmentation.
    }\label{fig:entropy}
\end{figure*}

In summary, solving the above linear system gives the probability, for each pixel, to be associated with a given seed. These probabilities are typically fractional, which turns out to be a great advantage:  It allows the definition of meaningful spatially resolved uncertainty measures such as the entropy at each pixel, 
\begin{equation}\label{eq:Entropy}
H_v = - \sum_{a \in \mathcal{L}} Z_{v,a} \log \left( Z_{v,a} \right)
\end{equation}
See Figure~\ref{fig:entropy} for an illustration.

The same kind of measure would not be meaningful for graph cut type approaches, whose unimodularity guarantees integral all-or-nothing solutions. For graph cut results, the entropy would be constant across the entire image. 

In the Random Walker, the final labeling is obtained from the resulting assignment matrix by a winner-take-all selection of the 
the label with maximal probability for each vertex $v \longmapsto \argmax_{a \in \mathcal{L}} Z_{v, a}$.

\subsection{Structured Learning of a CNN to Predict Edge Weights for the Random Walker}

To find appropriate edge weights, early work~\cite{Grady06} used simple parametric estimators of local image
contrast, such as exponentiated intensity differences.
Later work has used flexible nonlinear estimators, such as random forests,
to estimate edge weights~\cite{Jancsary12}.
We want to use gradient descent to train weight estimators in a structured fashion in the sense of~\cite{tappen07}, only for seeded segmentation and for a more flexible and powerful class of estimators.

When solving equation~\eqref{eq:RandomWalkerLinearSystem} exactly, structured learning amounts to the following optimization problem: the minimization of loss $l$
with respect to the neural network parameters $\Theta$
\begin{alignat}{1}
    \argmin_{\Theta} l\left(Z_U^*, -L_U(I;\Theta)^{-1}
    \cdot B(I;\Theta)^T \cdot Z_M\right)
\end{alignat}
based on  ground truth $Z_U^*$. Here, we have made explicit the dependence of $L_U$ and $B^T$ on the edge weights $w$, which in turn depend
on image $I$ and network parameters $\Theta$.

To solve the optimization problem we use gradient descent,
for which we compute the gradient
\begin{alignat}{1}
    \frac{\partial l\left(Z_U^*, Z_U\right)}{\partial \Theta}
    = \frac{\partial l\left(Z_U^*, Z_U\right)}{\partial Z_U}
    \frac{\partial Z_U}{\partial w}
    \frac{\partial w}{\partial \Theta}.
\label{eq:dldtheta}
\end{alignat}
The first and third term on the right hand side are standard: the partial derivative of the loss function with respect to a candidate solution, and of the neural network with respect to its parameters, respectively.  The remaining tensor is cumbersome due to its extremely large dimensions: the partial derivative of the probability of
(typically:) dozens of seeds $|\mathcal{L}|$ in millions of pixels $|U|$ with respect to millions of edges $|E|$ make for a formidable object
$\partial Z_U / \partial w \in \mathbb{R}^{|U| \times |\mathcal{L}| \times |E|}$.
One way to evaluate this expression is to take the derivative with respect
to $w$ of the linear system~\eqref{eq:RandomWalkerLinearSystem}:
\begin{equation*}
    \frac{\partial L_U Z_U}{\partial w} = -\frac{\partial B^T Z_M}{\partial w}
\end{equation*}
Since the probabilities at the marked pixels do not depend on the edge
weights we have $\partial Z_M / \partial w = 0$ and obtain with the product rule
the following tensor-valued linear system, whose solution is $\partial Z_U / \partial w$:
\begin{alignat}{1}\label{eq:GradientLinearSystem}
	L_U \frac{\partial Z_U}{\partial w} =
	- \frac{\partial L_U}{\partial w} Z_{U} -\frac{\partial B^T}{\partial w} Z_M
\end{alignat}
This equation is a combined representation of $|\mathcal{L}||E|$ usual linear
systems with matrix $L_U$ and changing right hand sides.
These right hand sides are easy to calculate as
$\partial L_U / \partial w$ and $\partial B^T / \partial w$
are very sparse and constant with respect to $w$, i.e.\ they do not change during
the gradient descent.

\subsection{Simplifications for Calculating the Gradient}
\label{sec:simplifications}
On the other hand, computing the huge rank 3 tensor $\partial Z_U / \partial w$ requires,
in each gradient descent step, solving the tensor-valued linear system~\eqref{eq:GradientLinearSystem},
which is computationally expensive.
As the tensor is only used in the tensor multiplication
\begin{alignat*}{1}
    \frac{\partial l(Z_U^*, Z_U)}{\partial w} = \frac{\partial l(Z_U^*, Z_U)}{\partial Z_U} \frac{\partial Z_U}{\partial w} \in \mathbb{R}^{|\mathcal{L}|}
\end{alignat*}
we can make a few simplifying approximations:

\paragraph{Sparse Gradient}
Instead of calculating the entire gradient tensor $\partial l(Z_U^*, Z_U) / \partial w$
we randomly select $n \ll |E|$ edges for which we solve the corresponding linear
systems and set the entries corresponding to other edges to zero.
This approach can be seen as a stochastic gradient descent algorithm.

We have tried more sophisticated strategies, including ones based on the label entropy or ones that concentrate on misclassified pixels.
However, none of them performed significantly better than the simple and parameter-free uniform sampling of edges, so this was used in all experiments shown.

\paragraph{Gradient Pruning}

\begin{figure}
    \centering
    \begin{tikzpicture}
    \node (img)  {\includegraphics[width=0.4\textwidth]{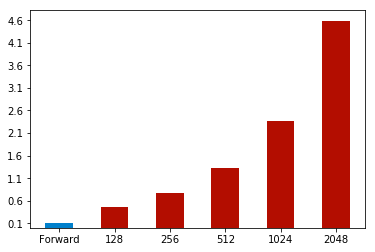}};
    \node[left of =  img, node distance=0cm, rotate=90, yshift=3.6cm, font=\color{black}] {Time ($s$)};
    \node[below of =  img, node distance=0cm, xshift=0.0cm, yshift=-2.55cm,font=\color{black}] {\# of Sampled Gradients};
    \end{tikzpicture}
     \caption{Run time comparison for different number of gradients sampled. For comparison, a complete backpropagation step ($n=16384$) takes $37s$. All results reported are for a (i7-6700, 3.4 GHz) CPU machine.}\label{fig:time}
\end{figure}

In equation~\eqref{eq:dldtheta}, the entries in the huge 3-tensor are multiplied, and hence modulated, with entries from $\partial l(Z_U^*, Z_U)/\partial Z_U$. Inspired by \cite{Komodakis07},
for a given edge $(ij)$ we only compute the contributions from label
\begin{alignat*}{1}
  \argmax_a  \left| {\left( \frac{\partial l(Z_U^*, Z_U)}{\partial Z_U} \right)}_{i,a} \right|.
\end{alignat*}
Taken together, these two approximations reduce the size of the tensor
from $|U| \times |\mathcal{L}| \times |E|$ to $|U| \times 1 \times n$, i.e
instead of solving $|\mathcal{L}| |E|$ linear systems of size $|U|$ we only
solve $n$ linear systems of size $|U|$.
As an example, we choose a 4-connected graph with $128 \times 128$ vertices and $10$ labels. Fig~\ref{fig:time} shows average backpropagation run-times. We found $n=1024$ to be the best compromise between ($\approx 15$-fold) speed-up and accuracy.

\section{Experiments and Results}
\label{sec:experiments}
\subsection{Pipeline Overview}

The proposed method is agnostic to CNN architecture; we employ a Convolutional-Deconvolutional network inspired by~\cite{Lee17}.
The segmentation is conducted in 2D, but the network accepts a 3D volume as input (3 slices along z) and predicts edge weights for the central slice only.
In our implementation we used the 4-connected graph, so on average we only need two edge weights per pixel, one horizontal and one vertical.
The network details are shown in the supplementary material.
The network outputs edge weight images at half the original resolution.
This speeds up both inference and backpropagation and reduces the memory footprint without impacting the final result.
The assignments are then scaled to the original size using bilinear interpolation before computing the loss.

As loss function we choose the Cross Entropy loss (CE) on the assignment matrix, defined as
\begin{equation}
	\text{CE}(Z^*, Z) = -
	\frac{1}{|V|}
	\sum_{i \in V}
	\sum_{a \in \mathcal{L}}
	Z^*_{i, a} \log\left(Z_{i, a}\right)
\end{equation}
where $Z$ is our calculated assignment matrix and $Z^*$ the ground truth.
We also tried employing the Mean Squared Loss and the Dice Loss.
While results are comparable, CE performed slightly better.
In addition, we used an unstructured CE loss on the weights and for regularization an $\ell^2$ weight decay on
the network parameters $\Theta$.
In summary, the loss reads
\begin{alignat}{2}
	l\left(Z^*, Z(\Theta) \right) =
	\text{CE}(Z^*, Z(\Theta))
	+ \alpha \text{CE}\left(w^*, w(\Theta)\right) \\ \nonumber
	+ \frac \gamma 2 \left\lVert \Theta \right\rVert_2^2.
\end{alignat}
where $w^*$ are the ground truth edges obtained from $Z^*$, $\alpha=10^{-2}$ and $\gamma=10^{-5}$.

The network is trained on patches of size $256 \times 256$.
We use mini-batches of size $10$ and train for a total of $10$ epochs.
Before the structured training, the CNN is pre-trained using only the side loss
on the same data, with mini-batches of size $10$ and for $2$ epochs.
As optimizer we use Adam~\cite{ADAM}.

\subsection{Seeded CREMI 2D Segmentation}

\label{Cremi 2D segmentation}
In our experiments we work on the data from the MICCAI Challenge on Circuit Reconstruction from Electron Microscopy Images (CREMI)~\cite{CREMI}. The challenge has been designed to measure automated segmentation accuracy in connectomics. There, the aim is to trace a myriad neural processes in serial section electron microscopy images to ultimately produce a wiring diagram of the brain. While the accuracy of automated methods is improving continuously \cite{Funke2018}, they do not yet reach a precision that would allow immediate downstream processing. Also, automated methods need much training data, and so seeded segmentation of connectomics data is of interest both for ground truth generation and for proofreading of automated predictions.

The CREMI dataset is composed of three volumes from adult Drosophila melanogaster
(common fruit fly) brain tissue, each of size $1250\times1250\times125$ pixels.
The three volumes are very different in appearence: 
While the neurites in CREMI A are mostly homogeneous in size and shape,
the two other volumes (CREMI B and C) are more challenging, with cells that have
jagged boundaries and large variations in size and shape.
We use the first 50 slices in $z$ for testing and the last 75 for training.

For our seeded segmentation, we assume that an oracle provides precisely one seed per segment.
Here we implement an oracle by automatically computing the seeds from the ground truth.
For each segment we place one seed at a random location, but with a reasonable
distance to any boundary.
Because using seeds from an oracle is a strong assumption we cannot directly compare our
approach to the unseeded algorithms competing in the CREMI challenge.
Therefore, we evaluate the performance of our end-to-end algorithm by comparing
it to the following pipelines, which are also seeded approaches:
\paragraph{Standard Random Walker:}
We slightly modified our network to directly predict boundary probability maps.
For this, we trained the CNN on the same dataset for a total of $10$ epochs,
while using the Dice loss and mini-batches of size $10$.
Subsequently, we compute the segmentation using the standard Random Walker algorithm given in~\cite{Grady06}.
The algorithm has a single hyperparameter $\beta$, which we tune optimally.
As for the Learned Random Walker we downsampled the graph by a factor of 2 to reduce computational footprint.
\paragraph{Watershed:}
For the Watershed algorithm, we used the same methodology as for the standard Random Walker algorithm to predict boundary probability maps.
The only difference is in the output size. Indeed, the Watershed algorithm is very efficient to compute and thus we do not downsample.
\paragraph{Learned Watershed:}
Lastly, we compared our results with the numbers published in~\cite{LWS}.

All segmentations are evaluated using the official CREMI metrics: Variation of Information (VOI) and Adapted Rand Error (ARAND).
VOI~\cite{Meila05} is the conditional entropy of the predicted and the ground truth segmentation:
$\text{VOI} =  \text{VOI}_{\text{split}} + \text{VOI}_{\text{merge}} = H(Z | Z^*) + H(Z^* | Z)$
where $H$ is the conditional entropy.
ARAND is the complement of the Adjusted Rand Index~\cite{Rand71}:
$\text{ARAND} = 1 - \text{AdjRAND}.$
According to the challenge, we used a tolerance of two pixels around the boundary.

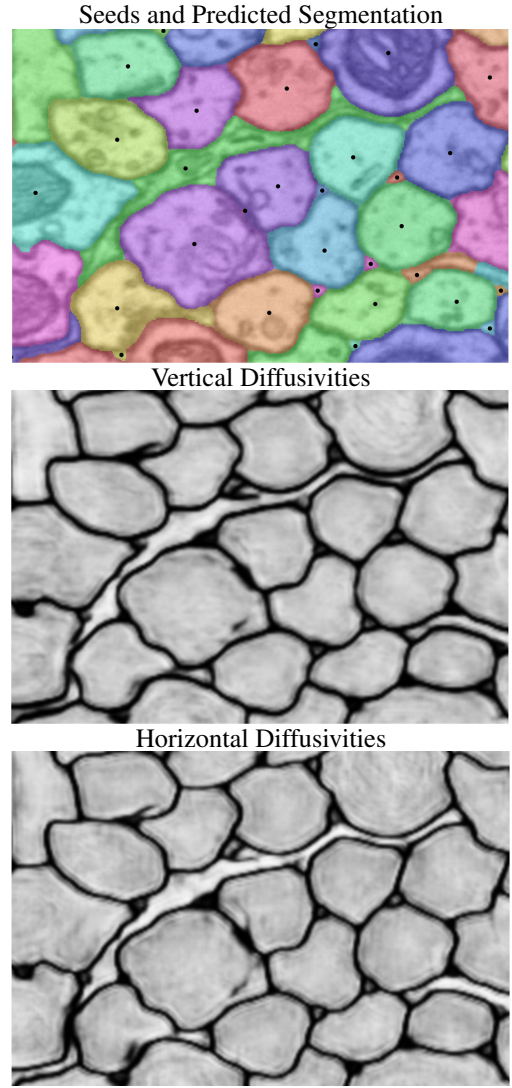
\begin{figure}
    \centering
    \begin{tikzpicture}
    \node (img)  {\includegraphics[width=0.38\textwidth]{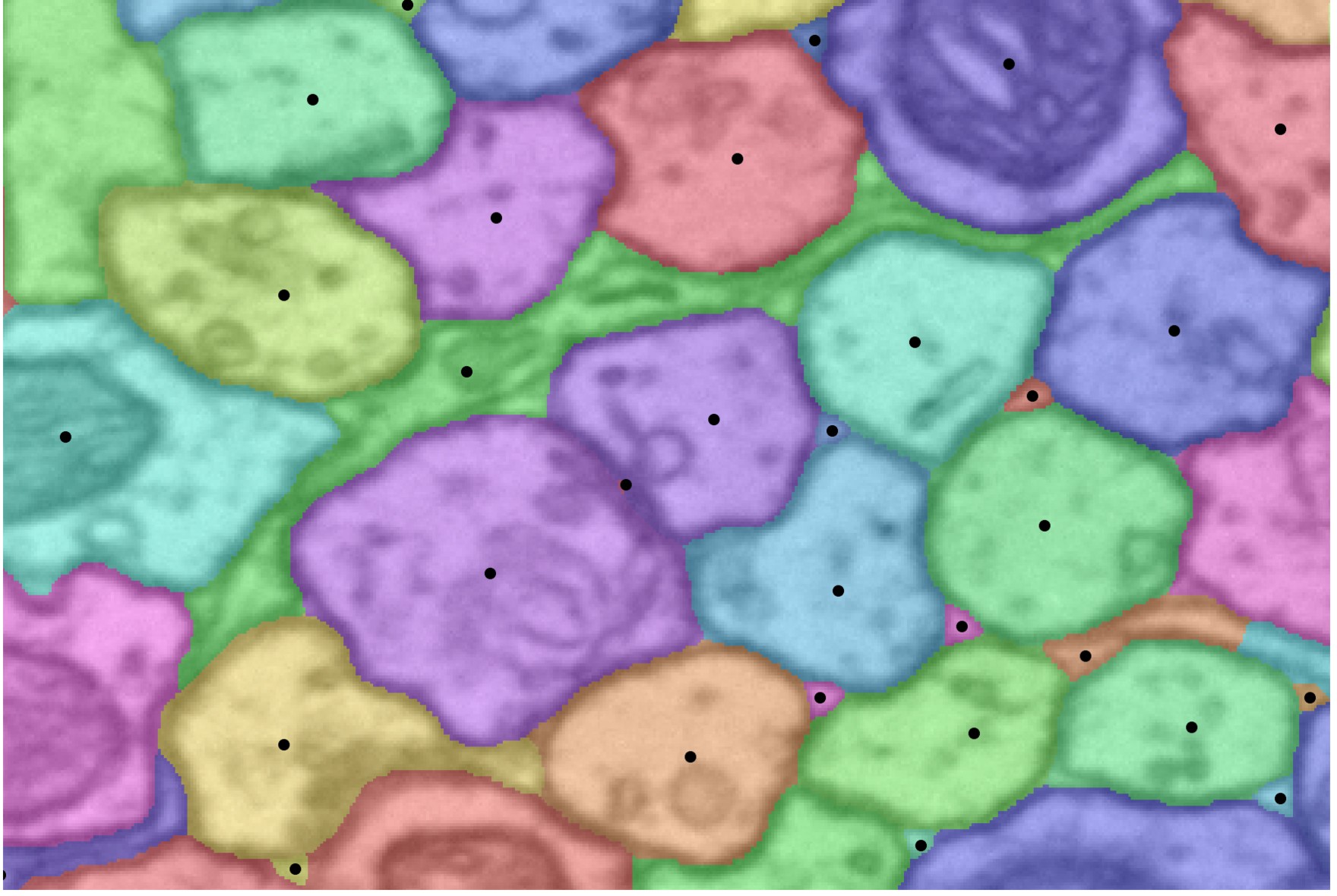}};
    \node (img2)[below of = img, yshift=-3.8cm] {\includegraphics[width=0.38\textwidth]{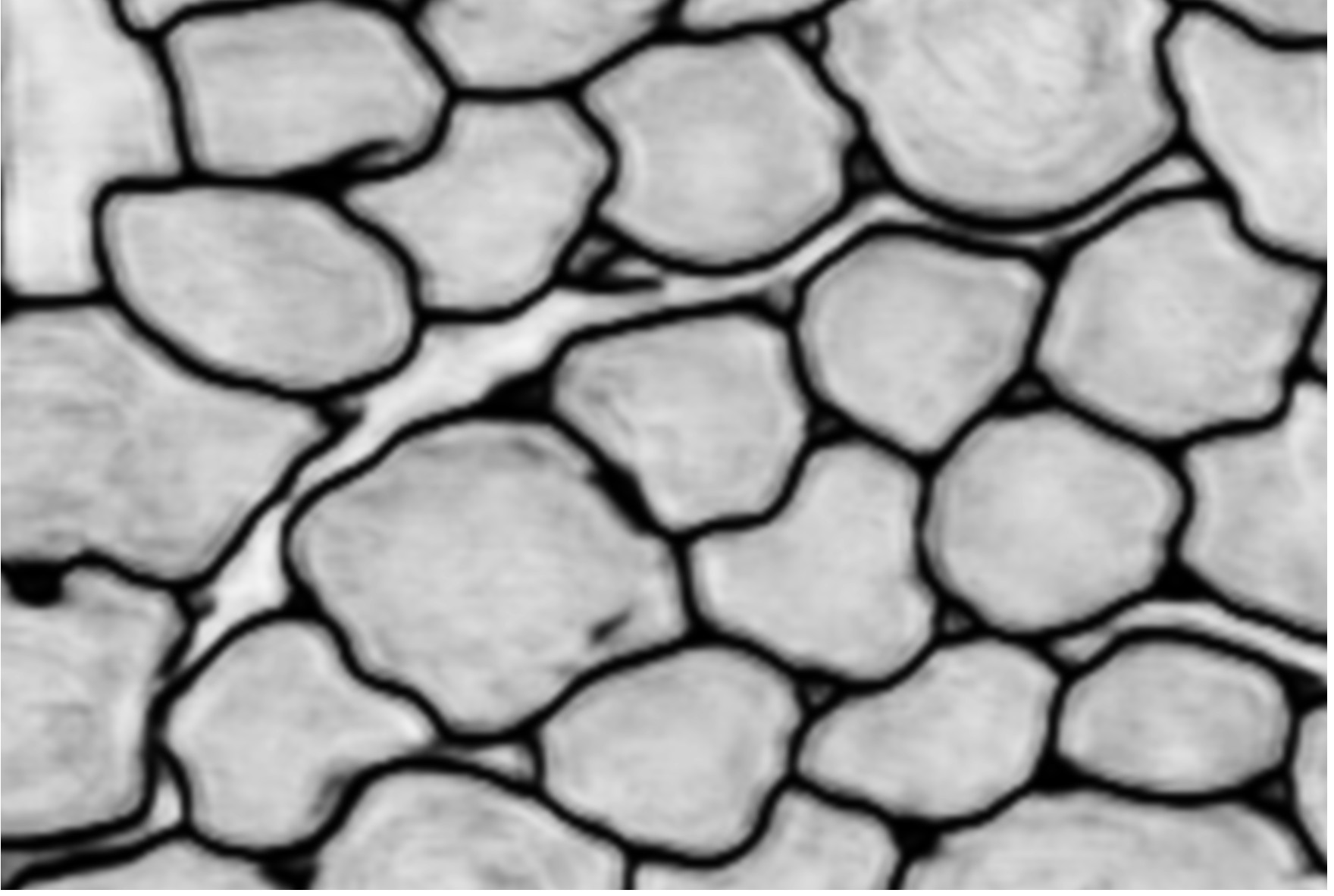}};
    \node (img3)[below of = img2, yshift=-3.8cm] {\includegraphics[width=0.38\textwidth]{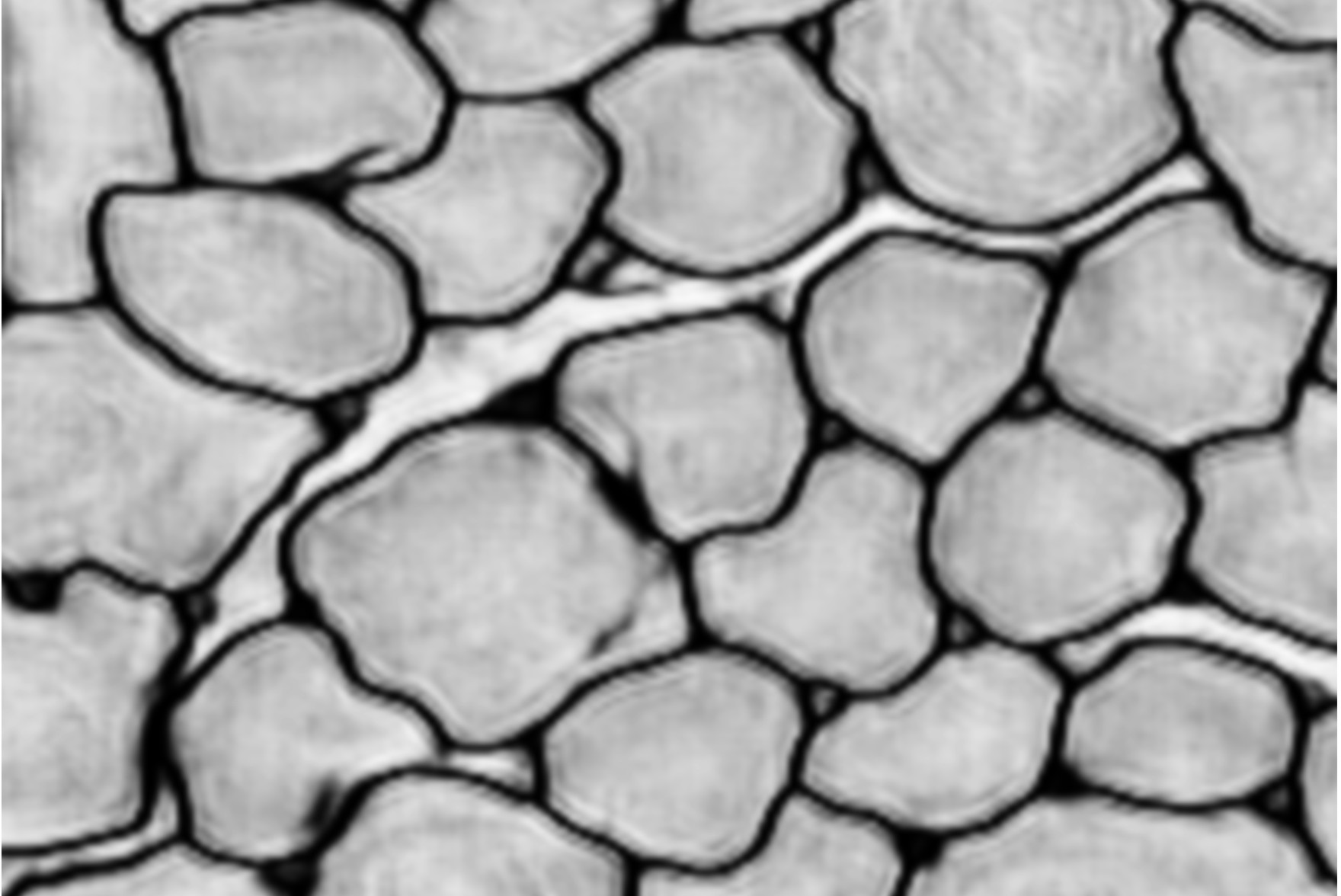}};
    
    \node[above of =  img, node distance=0cm, xshift=0.0cm, yshift=2.4cm,font=\color{black}] {Seeds and Predicted Segmentation};
    \node[above of =  img2, node distance=0cm, xshift=0.0cm, yshift=2.4cm, font=\color{black}] {Vertical Diffusivities};
    \node[above of =  img3, node distance=0cm, xshift=0.0cm, yshift=2.4cm,
    font=\color{black}] {Horizontal Diffusivities};
    \end{tikzpicture}
     \caption{Learned Random Walker qualitative behavior in narrow corridors. Top: seeds and the resulting Learned Random Walker segmentation. Middle and bottom:  
    The learned horizontal and vertical diffusivities
    are much stronger inside a corridor than near its perimeter.
    The horizontal diffusivity (bottom image) is a little larger because the corridor itself is horizontal. 
}\label{fig:res_cremi3}
\end{figure}

\begin{table*}[h]
    \begin{center}
	\centering
	\begin{tabular}{c c c c c c}
    \hline
       VOI & WS & LWS & RW  & LRW\\
     \hline\hline
     Cremi A  & $0.075 \pm 0.024$ & ---& $0.177 \pm 0.015 $ &$\textbf{0.062} \pm \textbf{0.021}$ \\
     Cremi B & $0.211 \pm 0.080$ & --- & $0.362 \pm 0.086 $ &$\textbf{0.193} \pm \textbf{0.089}$\\
     Cremi C & $\textbf{0.209} \pm \textbf{0.074}$ & --- & $0.421 \pm 0.091 $ &$0.232 \pm 0.081 $ \\
     Total       & $0.165 \pm 0.091$ & $0.376 \pm 0.034$& $0.320 \pm 0.127 $&$\textbf{0.162} \pm \textbf{0.102}$ \\
     \end{tabular}

	\begin{tabular}{c c c c c c}
        \hline
        ARAND & WS & LWS & RW  & LRW\\
        \hline\hline
        Cremi A & $0.016 \pm 0.010$ & --- & $0.042 \pm 0.008$ &$\textbf{0.011} \pm \textbf{0.009}$\\
        Cremi B & $0.049 \pm 0.044$ & --- & $0.153 \pm 0.078$ &$\textbf{0.045} \pm \textbf{0.044}$ \\
        Cremi C & $\textbf{0.053} \pm \textbf{0.045}$ & --- & $0.163 \pm 0.066$ &$0.061 \pm 0.038$ \\
        Total       & $\textbf{0.039} \pm \textbf{0.037}$ & $0.082 \pm 0.001$& $0.239 \pm 0.146 $ &$\textbf{0.039} \pm \textbf{0.040}$\\
        \hline
    \end{tabular}
    \vspace{0.23cm}
        \caption{Quantitative comparison of Seeded Watershed on a good boundary probability
        map, Learned Watershed~\cite{LWS}, Random Walker on a good boundary
        probability map and Learned Random Walker
        on the seeded CREMI challenge.
        Lower is better.}\label{tab:res_summary}
    \end{center}
\end{table*}

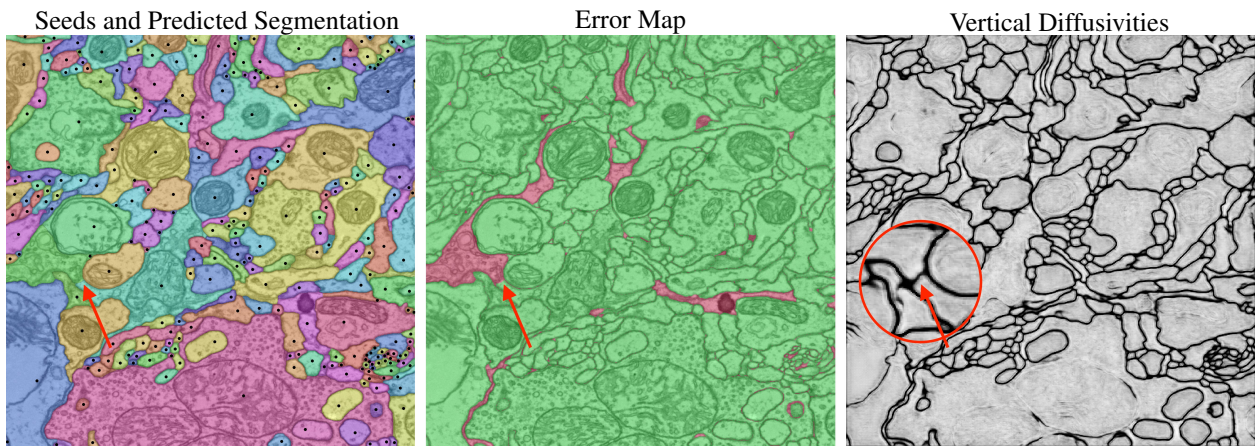
\begin{figure*}[t]
    \begin{center}
    \begin{tikzpicture}
    \node (img)  {\includegraphics[width=0.95\textwidth]{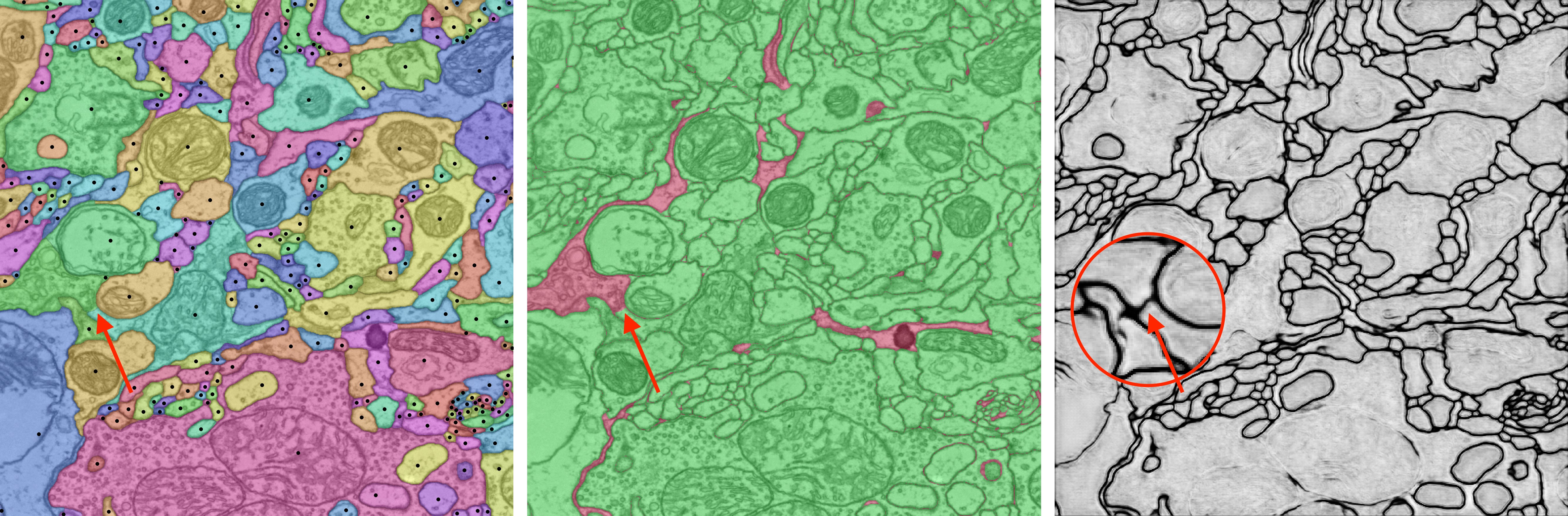}};
    \node[above of =  img, node distance=0cm, xshift=-5.5cm, yshift=+2.9cm,font=\color{black}] {Seeds and Predicted Segmentation};
    \node[above of =  img, node distance=0cm, xshift=0.0cm, yshift=+2.9cm, font=\color{black}] {Error Map};
    \node[above of =  img, node distance=0cm, xshift=5.7cm, yshift=+2.9cm,font=\color{black}] {Vertical Diffusivities};
    \end{tikzpicture}
     \caption{Typical Learned Random Walker fail on CREMI C.
    The red circle in the right image indicates a false positive boundary
    detection.
In this part of the data, the Learned Random Walker tends to hallucinate boundaries. 
    }\label{fig:res_cremi2}
    \end{center}
\end{figure*}

\subsection{Seeded CREMI 2D Segmentation Results}
\label{sec:results}
We show all results in Table~\ref{tab:res_summary}. Quantitatively, the Learned Random Walker with its structured training outperformed the Random Walker algorithm with unstructured training in every experiment.
Furthermore, the Learned Random Walker gave the best results in two of the three volumes, CREMI A and B, whereas Watershed obtained the best results on CREMI C.

Indeed, volumes B and C are less well
suited for the Random Walker algorithm, which has intrinsic shrinkage bias, whereas
watershed does not~\cite{Grady07}.
We observe that the structured learning seems to overcome this handicap:
While the standard Random Walker algorithm tends to suffer in narrow
corridors or near narrow funnels, the structured training helped the network predict stronger diffusivities inside this kind of regions, see Figure~\ref{fig:res_cremi3}.
We can assume the network learns a strong shape prior on how these kind of regions look like.

On the other hand, we observed a tendency to false positive boundary hallucination
with the pure structured training of the network.
An example is shown in Figure~\ref{fig:res_cremi2}.
Moreover, in our experiments, we observed that a CNN trained directly on a
boundary detection task generally performs better on more semantic tasks, like
distinguishing inner structures or mitochondria from true cell boundaries.

\subsection{Qualitative Experiments on Arabidopsis Thaliana Ovules}
\begin{figure}
    \begin{center}
    \centering
    \begin{tikzpicture}
    \node (img)  {\includegraphics[width=0.38\textwidth]{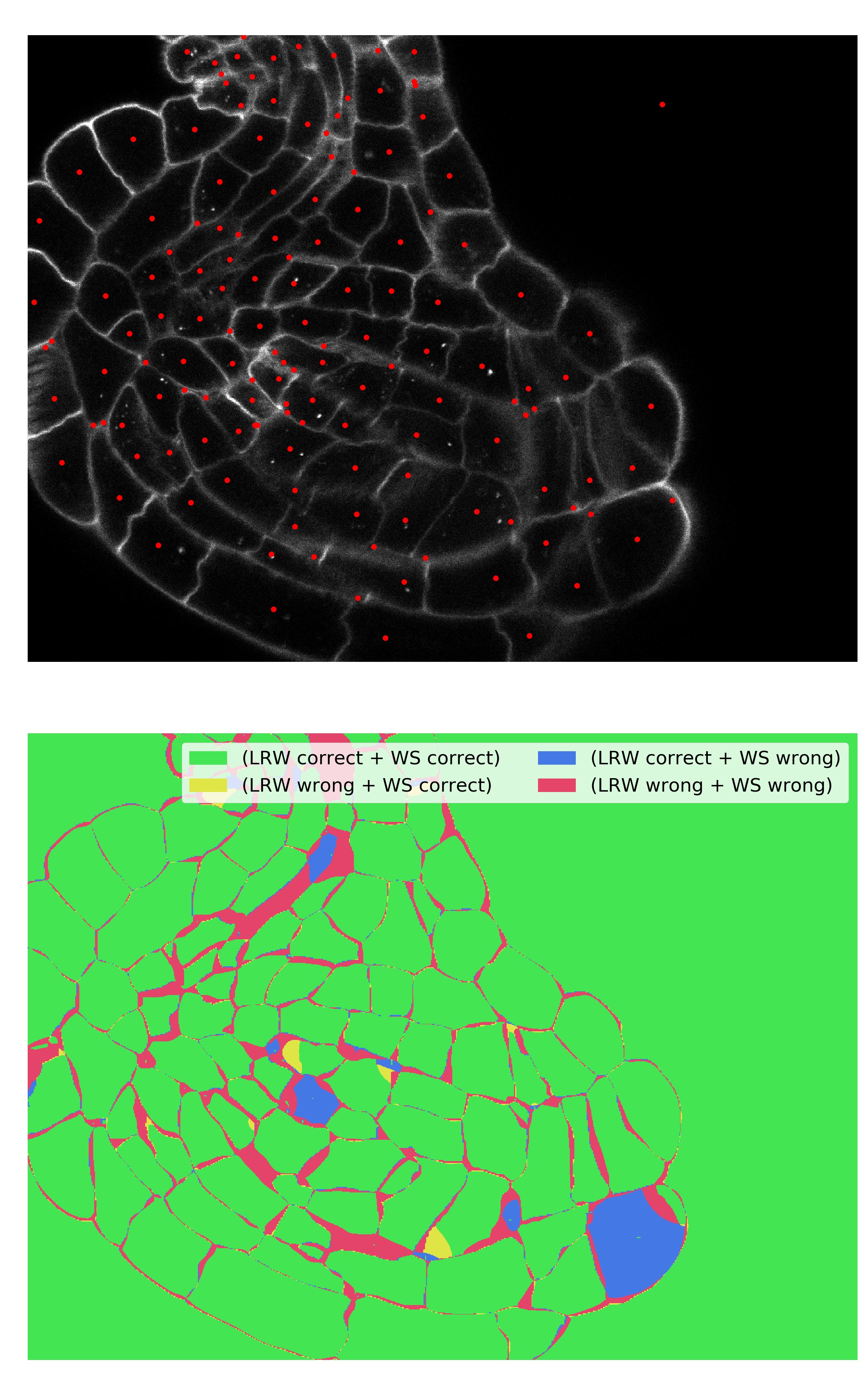}};
    \node[above of =  img, node distance=0cm, xshift=0cm, yshift=+5.3cm, font=\color{black}] {Seeds and Raw Input};
    \node[above of =  img, node distance=0cm, xshift=0cm, yshift=-0.05cm, font=\color{black}] {Error Map};
    \end{tikzpicture}
     \caption{Qualitative comparison between Learned Random Walker and Watershed applied on confocal microscopy data. Blue and yellow areas represent the regions where the Learned Random Walker outperformed Watershed and vice versa, respectively.
    }\label{fig:ovules}
    \end{center}
\end{figure}

In Figure~\ref{fig:ovules} we show a qualitative comparison between LRW and Watershed in a different imaging domain. The 3D dataset (courtesy of Schneitz Lab, TU Munich, Germany) consists of several Arabidopsis thaliana ovules, acquired using confocal laser scanning microscopy. The ovules are the major female reproductive organ of the plant. The Arabidopsis ovule serves as a prominent model for quantitative morphogenetic studies which require highly accurate segmentation of cells.
The experimental setup used is identical to sec.~\ref{sec:experiments}. As shown in Figure~\ref{fig:ovules}, the results agree qualitatively with our experiments on CREMI.

\subsection{Sampling Strategy vs.~Approximate Back-Propagation}
\label{sec:comparison}
Different approaches have been proposed for backpropagating gradient in
Laplacian systems. Very recently, a first order approximation of the true
derivative has been used in~\cite{Vernaza} for semantic segmentation.
Their approach has the conspicuous advantage that it requires solving only one system of linear equations.
We set up a simple experiment to show the differences between the Learned Random
Walker backpropagation and the one presented in~\cite{Vernaza}.

We tried to perfectly regress a single image labelling in two different scenarios: Extended seeding,
where we use large brush strokes on the image as seeds; and sparse seeds, where
each region was seeded in a single pixel only.
Instead of using a neural network to predict edge weights, we overparametrize the system by using one parameter per edge; and we
use gradient descent to update all these parameters until convergence.

For this example we used a sample image from the CREMI dataset~\cite{CREMI} and the 
same methodology as in section~\ref{sec:experiments}.
The quantitative results are presented in Table~\ref{tab:comparison}.
\begin{table}[h!]
    \begin{center}
    \centering
    \begin{tabular}{c c c}
        \hline
        ARAND & Extended & Sparse \\
        & Seeding & Seeding\\
        \hline\hline
        First order approx.~\cite{Vernaza} & 0.04 & 0.32\\
        LRW, 250 samples (ours) & 0.01 & 0.03\\
        LRW, no sampling & 0.01 & 0.01\\
        \hline
    \end{tabular}
    \vspace{0.23cm}
    \caption{Quantitative comparison between Learned Random Walker presented here
        and first order approximation from~\cite{Vernaza}.
        The ARAND metric is defined in Section~\ref{Cremi 2D segmentation}, and lower is better.
    }\label{tab:comparison}
    \end{center}
\end{table}

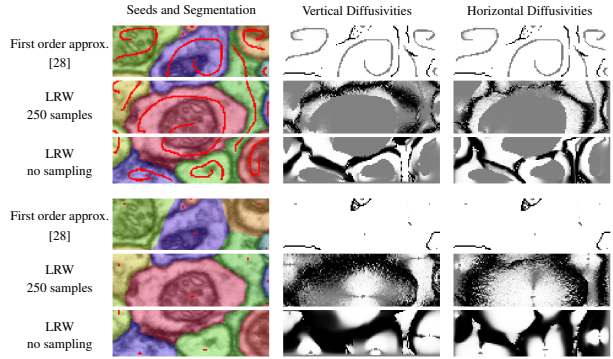
\begin{figure}
    \begin{center}
    \centering
    \begin{tikzpicture}
    \node (img)  {\includegraphics[width=0.4\textwidth]{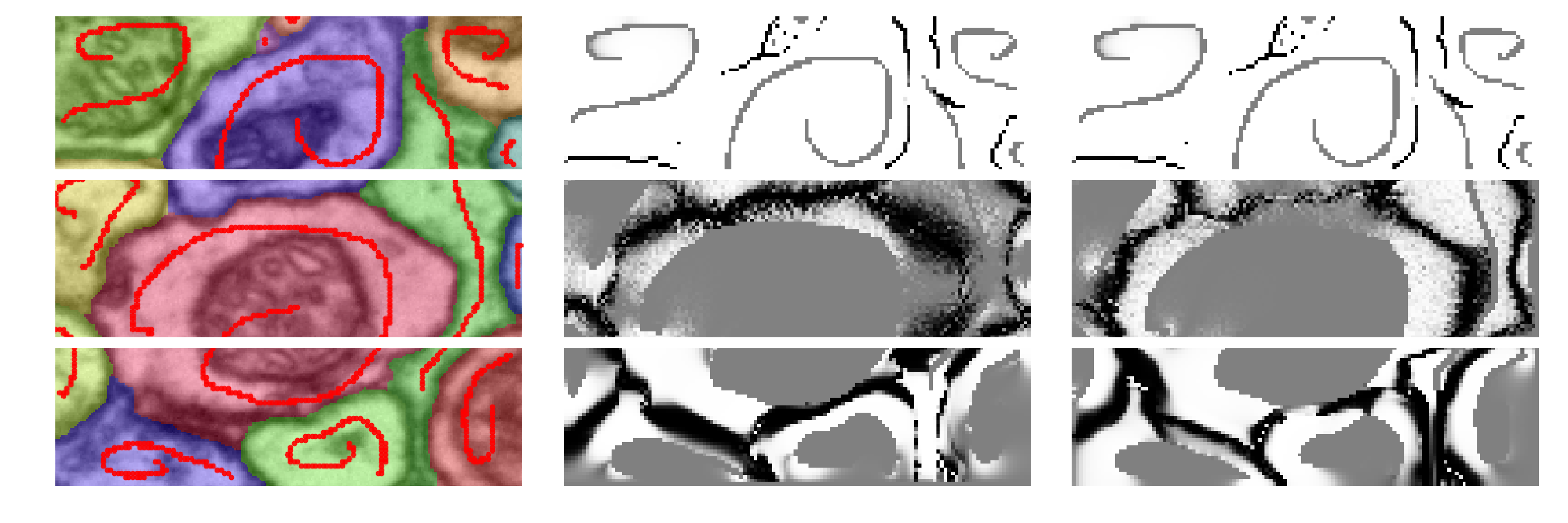}};
    \node[above of =  img, node distance=0cm, xshift=-2.2cm, yshift=+1.3cm, font=\color{black}] {\tiny{Seeds and Segmentation}};
    \node[above of =  img, node distance=0cm, xshift=0cm, yshift=+1.3cm, font=\color{black}]{\tiny{Vertical Diffusivities}};
    \node[above of =  img, node distance=0cm, xshift=2.3cm, yshift=+1.3cm, font=\color{black}] {\tiny{Horizontal Diffusivities}};
    
    \node[left of =  img, node distance=0cm, yshift=-0.6cm, xshift=-3.95cm, font=\color{black}] {\tiny{LRW}};
    \node[left of =  img, node distance=0cm, yshift=-0.85cm, xshift=-3.95cm, font=\color{black}] {\tiny{no sampling}};
    
    \node[left of =  img, node distance=0cm, yshift=+0.15cm, xshift=-3.95cm, font=\color{black}] {\tiny{LRW}};
    \node[left of =  img, node distance=0cm, yshift=-0.1cm, xshift=-3.95cm, font=\color{black}] {\tiny{250 samples}};
    
    \node[left of =  img, node distance=0cm, yshift=0.85cm, xshift=-3.95cm, font=\color{black}] {\tiny{First order approx.}};
    \node[left of =  img, node distance=0cm, yshift=0.6cm, xshift=-3.95cm, font=\color{black}] {\tiny{\cite{Vernaza}}};
    \node (img2) [below of = img, yshift=-1.3cm]{\includegraphics[width=0.4\textwidth]{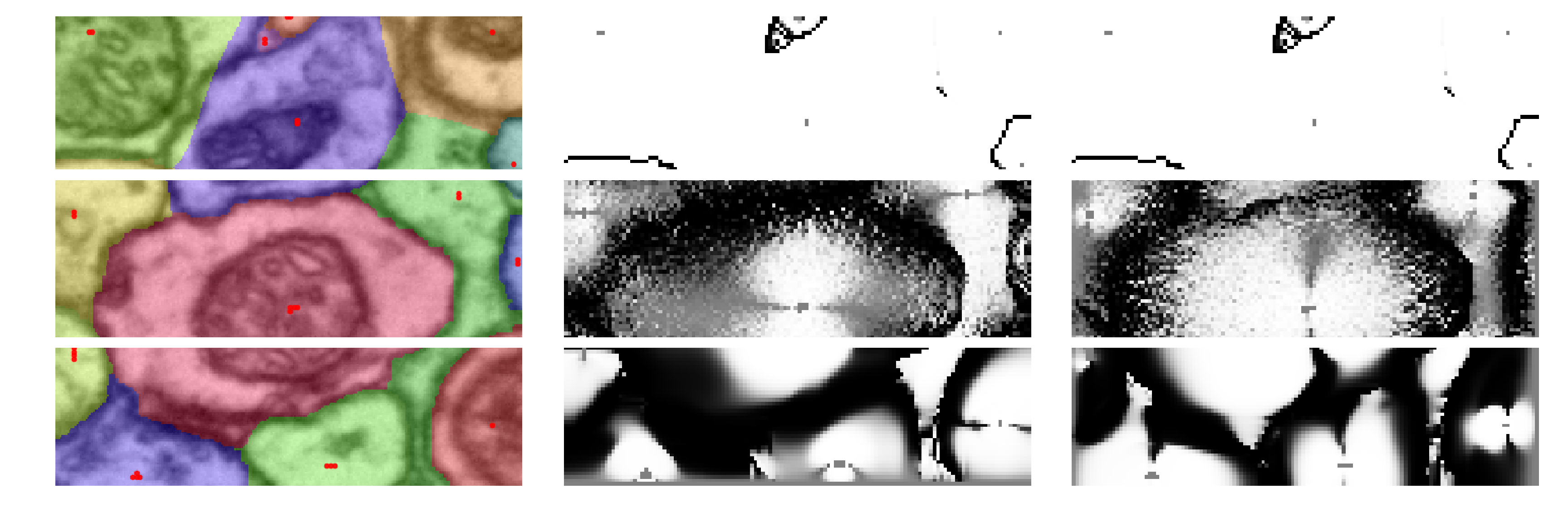}};
    \node[left of =  img2, node distance=0cm, yshift=-0.6cm, xshift=-3.95cm, font=\color{black}] {\tiny{LRW}};
    \node[left of =  img2, node distance=0cm, yshift=-0.85cm, xshift=-3.95cm, font=\color{black}] {\tiny{no sampling}};
    
    \node[left of =  img2, node distance=0cm, yshift=+0.15cm, xshift=-3.95cm, font=\color{black}] {\tiny{LRW}};
    \node[left of =  img2, node distance=0cm, yshift=-0.1cm, xshift=-3.95cm, font=\color{black}] {\tiny{250 samples}};
    
    \node[left of =  img2, node distance=0cm, yshift=0.85cm, xshift=-3.95cm, font=\color{black}] {\tiny{First order approx.}};
    \node[left of =  img2, node distance=0cm, yshift=0.6cm, xshift=-3.95cm, font=\color{black}] {\tiny{\cite{Vernaza}}};
    
    \end{tikzpicture}
     \caption{Qualitative comparison between Learned Random Walker presented here and first order approximation from~\cite{Vernaza}. We can observe that neither the sparse seeding nor our sampling strategy (sec.~\ref{sec:simplifications}) affect the reconstruction capability of the Learned Random Walker.
    }\label{fig:comparison}
    \end{center}
\end{figure}

We find that the Learned Random Walker can fit the ground truth both under extended and sparse seeds; whereas the  first-order approximation to backpropagation gives satisfactory results only with extended but not with sparse seeds. Qualitatively, we observed that the first-order approximation breaks down far from any given seeds Figure~\ref{fig:comparison}.\\
Moreover, we find that the sampling strategy introduced in sec.~\ref{sec:simplifications} has little effect on the  reconstruction accuracy.

\section{Acknowledgments}
We are grateful to Kay Schneitz, Rachele Tofanelli and Athul Vijayan for sharing data and helpful discussions.
This work was supported by the Deutsche Forschungsgemeinschaft (DFG) research unit 
FOR2581 Quantitative Plant Morphodynamics.

\section{Conclusion}
We have proposed an end-to-end learned pipeline for seeded segmentation.
We successfully trained a CNN jointly with the Random Walker algorithm obtaining
very competitive results and outperforming the standard Random Walker algorithm.
Furthermore, we propose and implemented an efficient sparse-backpropagation training and
experimentally proved that our method is able to train a network with very
sparse seeds.
In our experiments we always used a dense ground truth, but the proposed approach also allows
for training a network from sparse ground truth.
We plan to further explore this regime in future work.

{\small
\bibliographystyle{cvprstyle/ieee}
\bibliography{LRW}

\begin{thebibliography}{10}\itemsep=-1pt

\bibitem{Astroem2016}
Freddie \AA{}str\"om, Stefania Petra, Bernhard Schmitzer, and Christoph
  Schn\"orr.
\newblock {A} {G}eometric {A}pproach to {I}mage {L}abeling.
\newblock In {\em Proc.~ECCV}, 2016.

\bibitem{optnet}
Brandon Amos and J.~Zico Kolter.
\newblock {O}pt{N}et: Differentiable optimization as a layer in neural
  networks.
\newblock In Doina Precup and Yee~Whye Teh, editors, {\em Proceedings of the
  34th International Conference on Machine Learning}, volume~70 of {\em
  Proceedings of Machine Learning Research}, pages 136--145, International
  Convention Centre, Sydney, Australia, 06--11 Aug 2017. PMLR.

\bibitem{Sapiro07}
Xue Bai and Guillermo Sapiro.
\newblock A geodesic framework for fast interactive image and video
  segmentation and matting.
\newblock In {\em 2007 IEEE 11th International Conference on Computer Vision},
  pages 1--8, Oct 2007.

\bibitem{Baras10}
John Baras and George Theodorakopoulos.
\newblock {\em Path Problems in Networks}.
\newblock Synthesis Lectures on Communication Networks. Morgan {\&} Claypool
  Publishers, 2010.

\bibitem{Bertasius17}
Gedas Bertasius, Lorenzo Torresani, Stella~X. Yu, and Jianbo Shi.
\newblock Convolutional random walk networks for semantic image segmentation.
\newblock In {\em {CVPR}}, pages 6137--6145. {IEEE} Computer Society, 2017.

\bibitem{Boykov01}
Yuri~Y. Boykov and Marie-Pierre Jolly.
\newblock Interactive graph cuts for optimal boundary \& region segmentation of
  objects in n-d images.
\newblock In {\em Proceedings Eighth IEEE International Conference on Computer
  Vision}, volume~1, pages 105--112, 2001.

\bibitem{Chandra2016}
Siddhartha Chandra and Iasonas Kokkinos.
\newblock Fast, exact and multi-scale inference for semantic image segmentation
  with deep gaussian crfs.
\newblock In {\em Proc.~ECCV}, 2016.

\bibitem{chen2015}
Liang-Chieh Chen, Alexander Schwing, Alan Yuille, and Raquel Urtasun.
\newblock Learning deep structured models.
\newblock In {\em International Conference on Machine Learning}, pages
  1785--1794, 2015.

\bibitem{PWS}
Camille Couprie, Leo Grady, Laurent Najman, and Hugues Talbot.
\newblock Power watershed: A unifying graph-based optimization framework.
\newblock {\em IEEE Transactions on Pattern Analysis and Machine Intelligence},
  33(7):1384--1399, July 2011.

\bibitem{Cousty10}
Jean Cousty, Gilles Bertrand, Laurent Najman, and Michel Couprie.
\newblock Watershed cuts: Thinnings, shortest path forests, and topological
  watersheds.
\newblock {\em IEEE Transactions on Pattern Analysis and Machine Intelligence},
  32(5):925--939, May 2010.

\bibitem{CREMI}
CREMI.
\newblock Miccai challenge on circuit reconstruction from electron microscopy
  images, 2017.
\newblock \url{https://cremi.org}.

\bibitem{Falcao04}
A.~X. Falcao, J. Stolfi, and R. de Alencar~Lotufo.
\newblock The image foresting transform: theory, algorithms, and applications.
\newblock {\em IEEE Transactions on Pattern Analysis and Machine Intelligence},
  26(1):19--29, Jan 2004.

\bibitem{Funke2018}
Jan Funke, Fabian~David Tschopp, William Grisaitis, Arlo Sheridan, Chandan
  Singh, Stephan Saalfeld, and Srinivas~C Turaga.
\newblock Large scale image segmentation with structured loss based deep
  learning for connectome reconstruction.
\newblock {\em IEEE Transactions on Pattern Analysis and Machine Intelligence},
  2018.

\bibitem{Grady06}
Leo Grady.
\newblock Random walks for image segmentation.
\newblock {\em {IEEE} Trans. Pattern Anal. Mach. Intell.}, 28(11):1768--1783,
  2006.

\bibitem{Jancsary12}
Jeremy Jancsary, Sebastian Nowozin, and Carsten Rother.
\newblock Loss-specific training of non-parametric image restoration models: A
  new state of the art.
\newblock In {\em Proceedings of the 12th European Conference on Computer
  Vision - Volume Part VII}, ECCV'12, pages 112--125, Berlin, Heidelberg, 2012.
  Springer-Verlag.

\bibitem{ADAM}
Diederik~P. Kingma and Jimmy Ba.
\newblock Adam: {A} method for stochastic optimization.
\newblock {\em CoRR}, abs/1412.6980, 2014.

\bibitem{Komodakis07}
Nikos Komodakis and Georgios Tziritas.
\newblock Image completion using efficient belief propagation via priority
  scheduling and dynamic pruning.
\newblock {\em IEEE Transactions on Image Processing}, 16(11):2649--2661, Nov
  2007.

\bibitem{Lee17}
Kisuk Lee, Jonathan Zung, Peter Li, Viren Jain, and H.~Sebastian Seung.
\newblock Superhuman accuracy on the snemi3d connectomics challenge.
\newblock {\em CoRR}, abs/1706.00120, 2017.

\bibitem{Malmberg2012}
Filip Malmberg, Robin Strand, Joel Kullberg, Richard Nordenskj{\"o}ld, and
  Ewert Bengtsson.
\newblock Smart paint - a new interactive segmentation method applied to mr
  prostate segmentation.
\newblock MICCAI workshop, 2012.

\bibitem{Maninis18}
Kevis-Kokitsi Maninis, Jordi Pont-Tuset, Pablo Arbel{\'a}ez, and L.~Van Gool.
\newblock Convolutional oriented boundaries: From image segmentation to
  high-level tasks.
\newblock {\em IEEE Transactions on Pattern Analysis and Machine Intelligence},
  40(4):819--833, April 2018.

\bibitem{Meila05}
Marina Meil\u{a}.
\newblock Comparing clusterings: an axiomatic view.
\newblock In {\em In ICML 2005: Proceedings of the 22nd international
  conference on Machine learning}, pages 577--584. ACM Press, 2005.

\bibitem{Rand71}
William~M. Rand.
\newblock Objective criteria for the evaluation of clustering methods.
\newblock {\em Journal of the American Statistical Association},
  66(336):846--850, 1971.

\bibitem{Roth05}
Stefan Roth and Michael~J. Black.
\newblock Fields of experts: a framework for learning image priors.
\newblock In {\em 2005 IEEE Computer Society Conference on Computer Vision and
  Pattern Recognition (CVPR'05)}, volume~2, pages 860--867 vol. 2, June 2005.

\bibitem{tappen09}
Kegan G.~G. Samuel and Marshall~F. Tappen.
\newblock Learning optimized map estimates in continuously-valued mrf models.
\newblock In {\em 2009 IEEE Conference on Computer Vision and Pattern
  Recognition}, pages 477--484, June 2009.

\bibitem{Grady07}
Ali~Kemal Sinop and Leo Grady.
\newblock A seeded image segmentation framework unifying graph cuts and random
  walker which yields a new algorithm.
\newblock In {\em 2007 IEEE 11th International Conference on Computer Vision},
  pages 1--8, Oct 2007.

\bibitem{Spielman04}
Daniel~A. Spielman and Shang-Hua Teng.
\newblock Nearly-linear time algorithms for graph partitioning, graph
  sparsification, and solving linear systems.
\newblock In {\em Proceedings of the Thirty-sixth Annual ACM Symposium on
  Theory of Computing}, STOC '04, pages 81--90, New York, NY, USA, 2004. ACM.

\bibitem{tappen07}
Marshall~F. Tappen, Ce Liu, Edward~H. Adelson, and William~T. Freeman.
\newblock Learning gaussian conditional random fields for low-level vision.
\newblock In {\em 2007 IEEE Conference on Computer Vision and Pattern
  Recognition}, pages 1--8, June 2007.

\bibitem{Vernaza}
Paul Vernaza and Manmohan Chandraker.
\newblock Learning random-walk label propagation for weakly-supervised semantic
  segmentation.
\newblock In {\em 2017 IEEE Conference on Computer Vision and Pattern
  Recognition (CVPR)}, volume~00, pages 2953--2961, July 2017.

\bibitem{vonLuxburg07}
Ulrike von Luxburg.
\newblock A tutorial on spectral clustering.
\newblock {\em Statistics and Computing}, 17(4):395--416, Dec 2007.

\bibitem{LWS}
Steffen Wolf, Lukas Schott, Ullrich K{\"o}the, and Fred~A. Hamprecht.
\newblock Learned watershed: End-to-end learning of seeded segmentation.
\newblock {\em ICCV}, pages 2030--2038, 2017.

\bibitem{zheng2015}
Shuai Zheng, Sadeep Jayasumana, Bernardino Romera-Paredes, Vibhav Vineet,
  Zhizhong Su, Dalong Du, Chang Huang, and Philip~HS Torr.
\newblock Conditional random fields as recurrent neural networks.
\newblock In {\em Proceedings of the IEEE International Conference on Computer
  Vision}, pages 1529--1537, 2015.

\bibitem{Zhu03}
Xiaojin Zhu, Zoubin Ghahramani, and John Lafferty.
\newblock Semi-supervised learning using gaussian fields and harmonic
  functions.
\newblock In {\em Proceedings of the Twentieth International Conference on
  International Conference on Machine Learning}, ICML'03, pages 912--919. AAAI
  Press, 2003.

\end{thebibliography}
}
\clearpage
\appendix

\section{Neural Network Architecture}
\begin{figure}[h]
    \centering
    \includegraphics[width=0.45\textwidth]{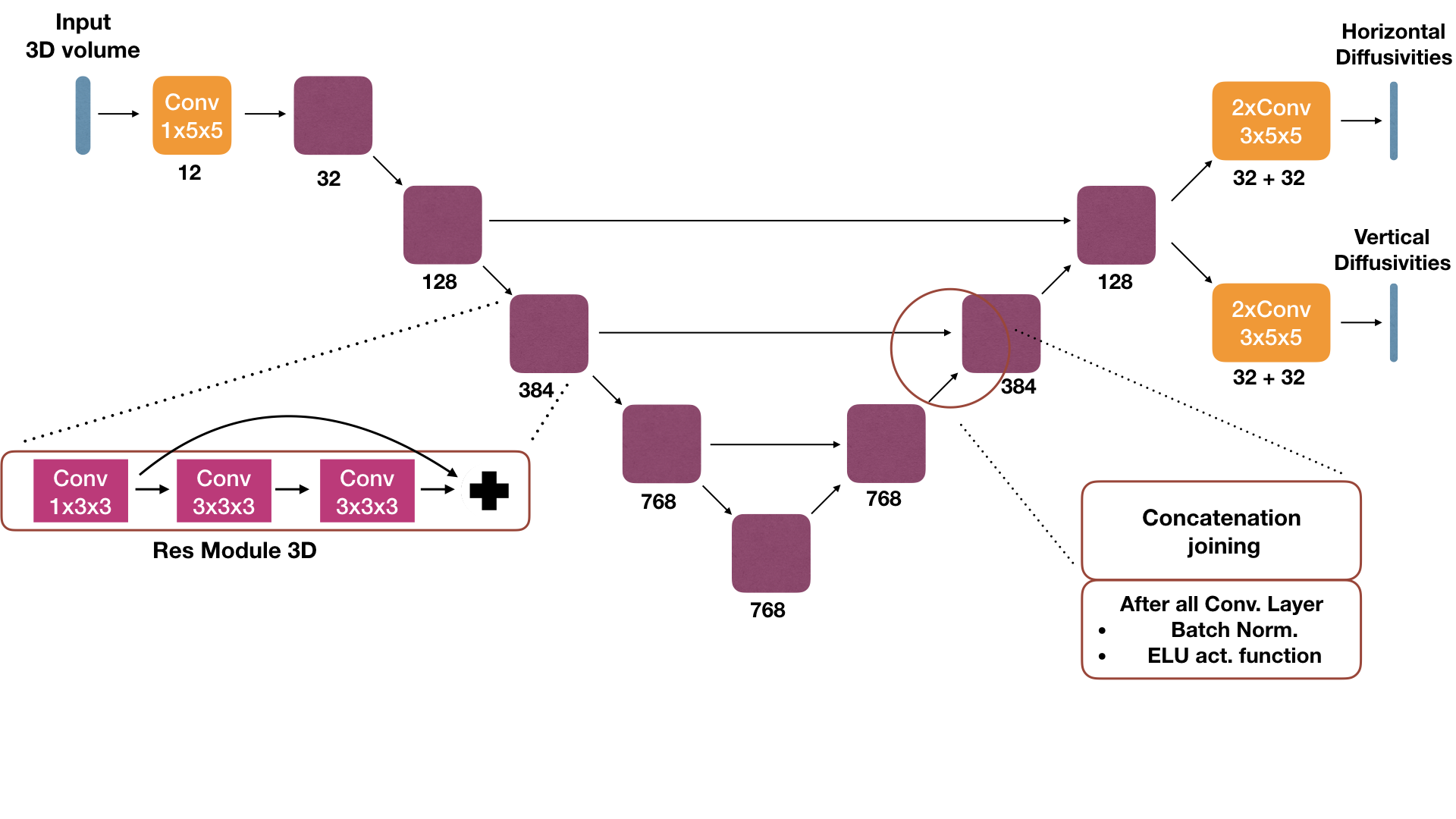}
    \caption{Illustration of the architecture used in our Learned Random Walker pipeline.}
    \label{fig:architecture_1}
\end{figure}

\begin{figure}[h]
    \centering
    \includegraphics[width=0.45\textwidth]{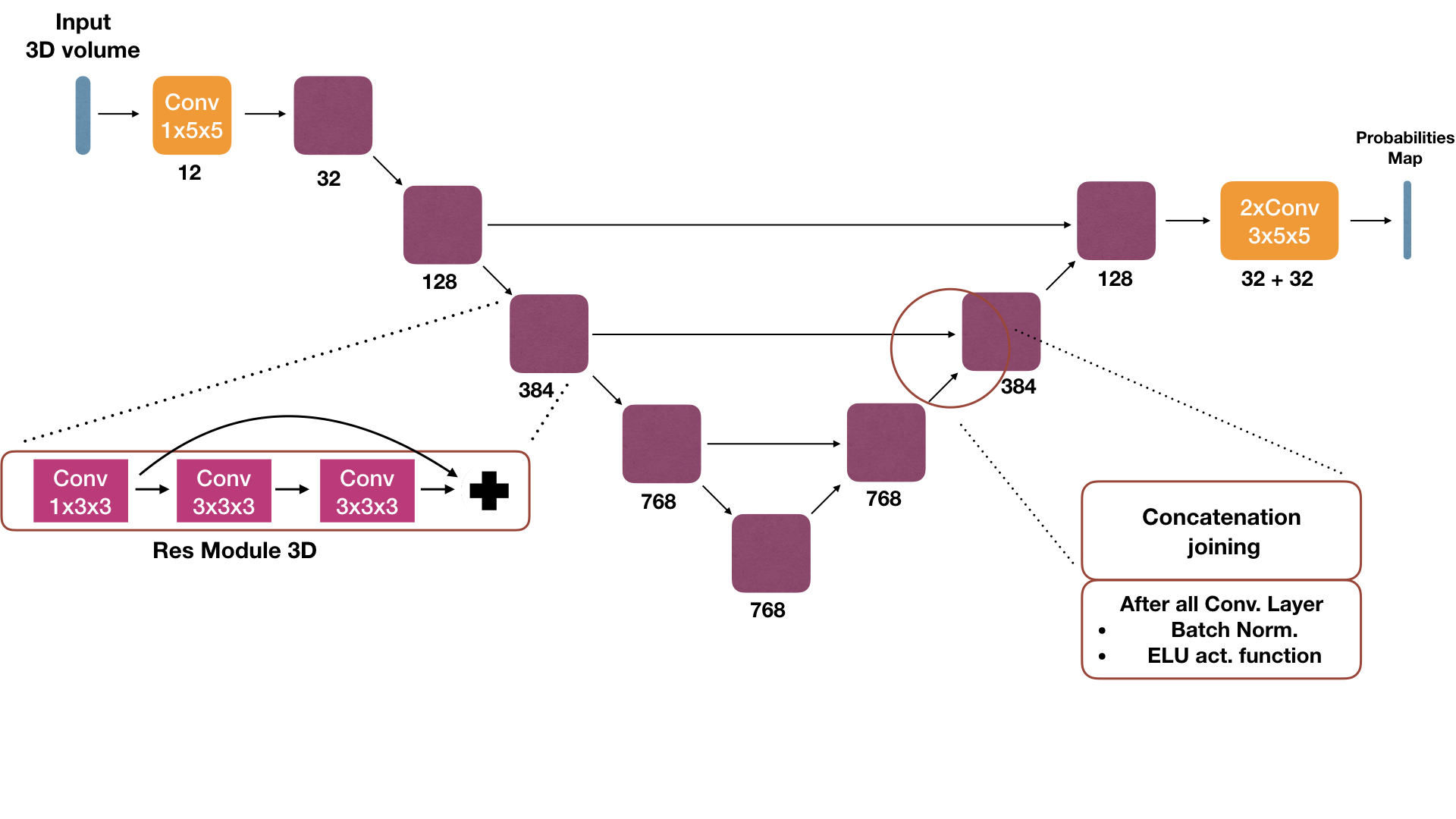}
    \caption{Illustration of the architecture used for our downsampled boundary probability map.}\label{fig:architecture_2}
\end{figure}

\begin{figure}[h]
    \centering
    \includegraphics[width=0.45\textwidth]{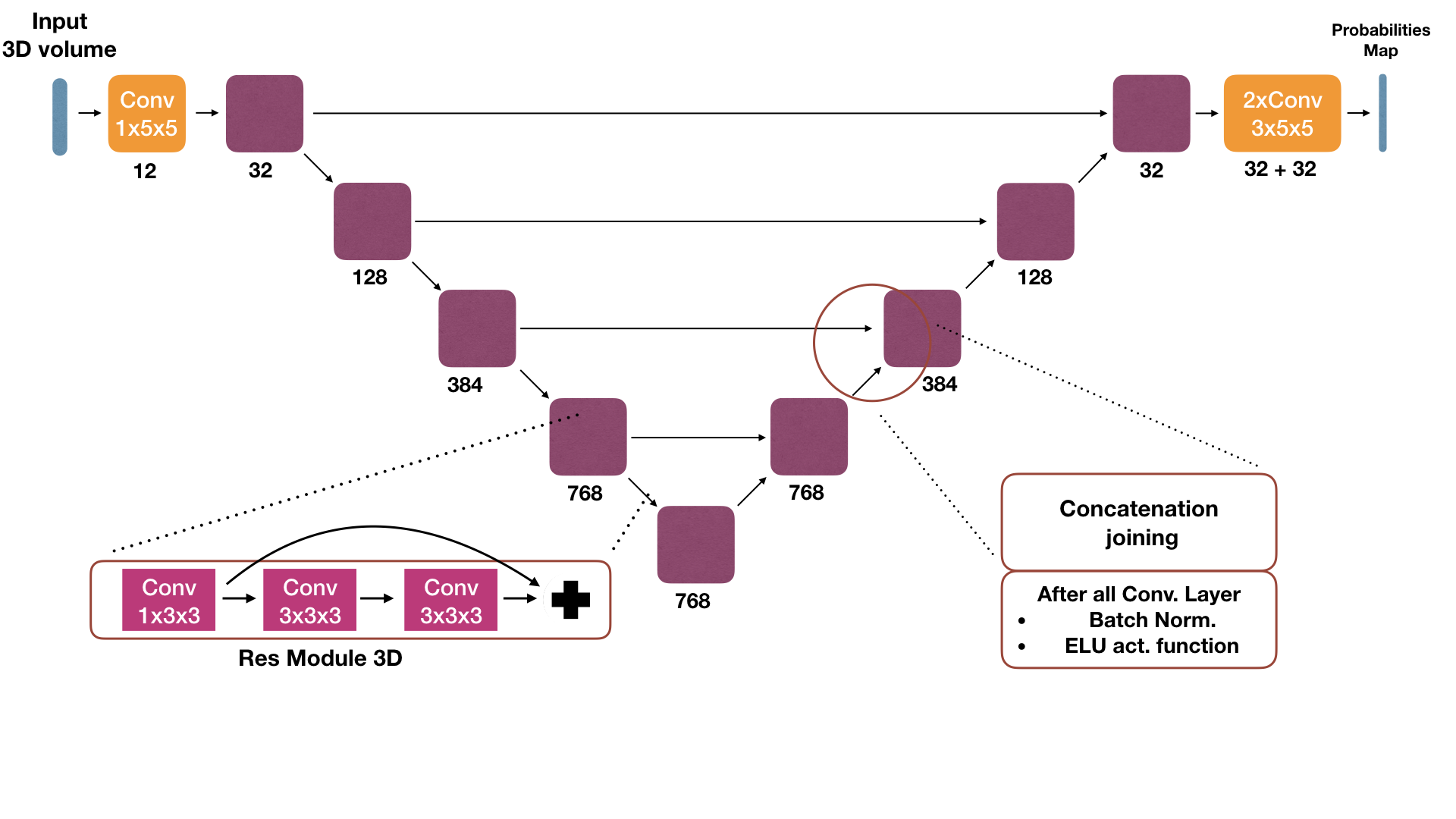}
    \caption{Illustration of the architecture used for our full size boundary probability map.}\label{fig:architecture_3}
\end{figure}

In this section we present a schematic description of the CNNs we used in our
approach (see Figures~\ref{fig:architecture_1},~\ref{fig:architecture_2} and~\ref{fig:architecture_3}).

All 3D convolutional layers use valid convolutions, except in the two output layers
where we removed the padding in the z-axis in order to project the volume back to 2D.
The filter size is written in each block and the number of filters below every block.

For downsampling (pointing down arrows), we used 3D maxpooling blocks with windows size of $1 \times 2 \times 2$.
For the upsampling (pointing up arrows), we used 3D transpose convolutions with filters size of $1 \times 2 \times 2$.

\section{Purely Structured Training}
In addition to the experiments we presented in the paper, we trained our Learned Random Walker pipeline without any side loss.
In its place, we used a log barrier on the edge weights as an unsupervised regularization.\\
With this, the new loss function reads:
\begin{alignat}{2}
		J\left(Z^*, Z, \Theta \right) =
	\text{CE}(Z^*, Z, \Theta)
	- \frac{\alpha }{2 |V|}
	\left\lVert \log(w) \right\rVert_1 \\ \nonumber
	+ \frac \beta 2 \left\lVert \Theta \right\rVert_2^2.
\end{alignat}

The terms are weighted by $\alpha=10^{-5}$ and $\beta=10^{-5}$.\\
The results obtained with this setup are presented in Table~\ref{tab:LRW_comparison}.
Despite the errors being larger without the side loss, the scores with the side loss are still competitive.

\begin{table}
	\centering
	\begin{tabular}{c c c}
 	\hline
        VOI & LRW with log barrier & LRW with side loss\\
  	\hline \hline
	CREMI A & $0.076 \pm 0.023$ &$0.062 \pm 0.021$\\
	CREMI B & $0.220 \pm 0.094$ &$0.193 \pm 0.089$\\
	CREMI C & $0.272 \pm 0.077$ &$0.232 \pm 0.081$\\
	Total   & $0.189 \pm 0.109$ &$0.162 \pm 0.102$\\
	\end{tabular}

	\begin{tabular}{c c c}
        \hline
        ARAND \ & LRW with log barrier & LRW  with side loss\\
        \hline \hline
        CREMI A & $0.014 \pm 0.077$ &$0.011 \pm 0.009$ \\
        CREMI B & $0.052 \pm 0.053$ &$0.045 \pm 0.044$ \\
        CREMI C & $0.067 \pm 0.036$ &$0.061 \pm 0.038$\\
        Total   & $0.044 \pm 0.043$ &$0.039 \pm 0.040$\\
        \hline
    \end{tabular}
    \vspace{0.23cm}
        \caption{Quantitative comparison of the Learned Random Walker with log barrier
            and with side loss by looking at the means and standard
            deviations over the test set.
    Lower is better.}\label{tab:LRW_comparison}
\end{table}

\end{document}